\documentclass[a4paper]{llncs}

\usepackage[utf8]{inputenc}
\usepackage{amssymb}
\usepackage{graphicx}
\usepackage[hidelinks]{hyperref}

\usepackage{url}
\newcommand{\keywords}[1]{\par\addvspace\baselineskip
\noindent\keywordname\enspace\ignorespaces#1}

\usepackage{cite}

\begin{document}

\mainmatter

\title{Context Spaces as the \\ Cornerstone of a Near-Transparent \& \\ Self-Reorganizing Semantic Desktop}

\author{
  Christian Jilek\inst{1,2} \and
  Markus Schröder\inst{1,2} \and
  Sven Schwarz\inst{1} \and \\
  Heiko Maus\inst{1} \and
  Andreas Dengel\inst{1,2}
}
\institute{
  Smart Data \& Knowledge Services Dept., DFKI GmbH, Kaiserslautern, Germany\\ \and
  Computer Science Dept., TU Kaiserslautern, Germany\\
  \email{\{firstname.lastname\}@dfki.de}
}
\maketitle

\begin{abstract}
Existing Semantic Desktops are still reproached for being too complicated to use or not scaling well.
Besides, a real "killer app" is still missing.
In this paper, we present a new prototype inspired by NEPOMUK and its successors having a semantic graph and ontologies as its basis.
In addition, we introduce the idea of context spaces that users can directly interact with and work on.
To make them available in all applications without further ado, the system is transparently integrated using mostly standard protocols complemented by a sidebar for advanced features.
By exploiting collected context information and applying Managed Forgetting features (like hiding, condensation or deletion), the system is able to dynamically reorganize itself, which also includes a kind of tidy-up-itself functionality.
We therefore expect it to be more scalable while providing new levels of user support.
An early prototype has been implemented and is presented in this demo.

\end{abstract}

\section{Introduction}
After its hype finally receded about half a decade ago, rather few advances in Semantic Desktop (SemDesk) research have been reported.
An overview of (modern) SemDesks can be found in \cite{DraganD12}:
Existing implementations are, for example, reproached for being rather complicated to use, not scaling well (thus draining lots of system resources), and there is still no real "killer app" available.
Concerning SemDesk applications, two categories could be observed:
newly created semantic applications and plug-ins to enhance traditional, non-semantic ones \cite{DraganD12}.

As a successor to the \textit{NEPOMUK Semantic Desktop}\footnote{\url{www.semanticdesktop.org}}, DFKI's Smart Data \& Knowledge Services department developed its own prototype\footnote{
meanwhile spanning over six years of permanent usage in the department and a group knowledge graph having approx. 2.6 million triple statements
} \cite{maus2013weaving} making SemDesk technology ready for 24/7 usage in practice, covering both, private and corporate scenarios.
After lessons learned in past\footnote{
e.g. \textit{ForgetIT} (\url{www.forgetit-project.eu}) and \textit{supSpaces} (\url{www.supspaces.de})
} and still ongoing projects\footnote{
e.g. \textit{Managed Forgetting} (\url{www.spp1921.de/projekte/p4.html.de})
}, we now propose \textit{Context Spaces} as an extension of this prototype addressing the issues mentioned before.

\section{Approach}
\begin{figure}
\centering
\includegraphics[width=1\textwidth]{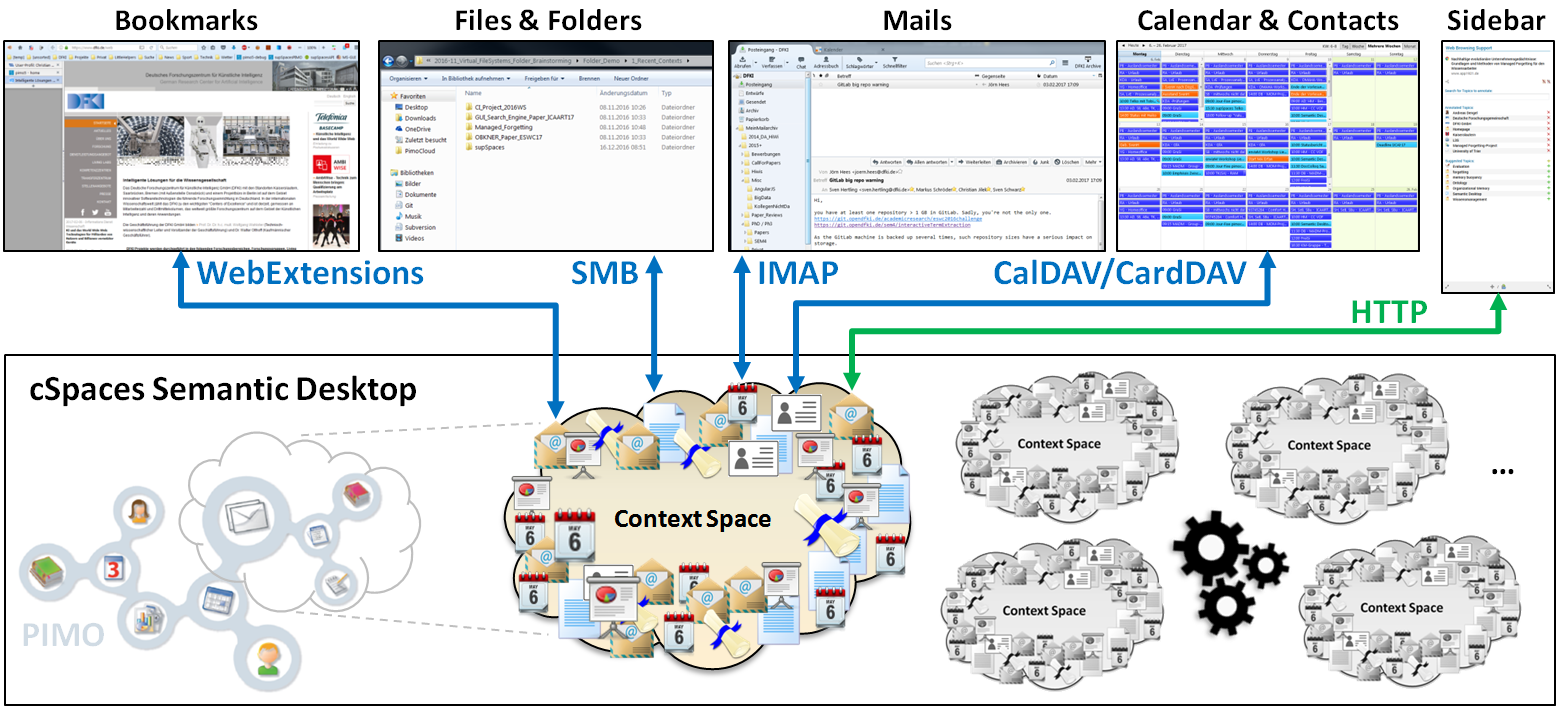}
\caption{Conceptual overview of the cSpaces Semantic Desktop}
\label{fig_cspaces}
\end{figure}

\noindent
\textbf{Context Spaces.}
One of SemDesk's cornerstones is the Personal Information Model (PIMO) \cite{sauermann2007pimo}, which tries to represent a user's mental model as good as possible.
Information items (files, mails, bookmarks, ...) that are related to each other in a person's mind, but are separated on their computer (file system, mail client, web browser, ...), can thus be interlinked.
With \textit{Context Spaces} (or \textit{cSpaces} for short) we extend this idea by explicitly (and additionally) associating items with contexts of the user (see lower left of Figure \ref{fig_cspaces}).
\vspace{-0.4cm}
\vspace{-0.4cm}
This is based on the intuition that every activity is performed in a certain context.
Hence, each information item stored on a person's computing device can be associated with one or more contexts (association strength may vary depending on the user's current context awareness).
We therefore assume that users are explicitly aware of the concept of context \cite{GomezPerez2009} and that they are also aware of their current context (at least most of the time).
Examples of such contexts are: \textit{Spain holiday 2017}, \textit{prepare ESWC18 paper}, or \textit{my childhood memories}.
We do not enforce a certain definition of context: users should be able to stick to their own conceptualization as much as possible.
However, we do assume that contexts express a certain relatedness of its elements.
Besides being a kind of container for things, they may also be strongly related to (calendar) events, tasks or cases.
Context hierarchies are also possible.
More details about our context model, which is an extension of \cite{SchwarzContextModel}, will be presented in another paper.
Instead of just having context as passive metadata, we in addition see it as an accessible element users can interact with (create new (sub)contexts, split or merge them, add/remove elements, etc.).

SemDesk user studies \cite{SauermannEval} revealed that people omitted rather specific relations in favor of basic ones (like \textit{isRelatedTo} or \textit{isPartOf}), whereas the system is formal where possible, e.g. representing calendar events or address book contacts.
This matches our idea of providing a low effort opportunity to already keep things a bit more tidied up when simply associating them with a certain context (or multiple).
Additionally, some of these associations may also be inferred by the system reducing manual effort even more, e.g. a received email reply can automatically be associated with the original mail's context.
More advanced features supporting the user will be discussed in the section after next.

\noindent
\textbf{Transparent Integration.}
Using contexts as an explicit interaction element only makes sense if applications also respect them.
Like illustrated in Figure \ref{fig_cspaces}, we therefore integrate cSpaces into the rest of the system using standard protocols like \textit{Server Message Block (SMB)} for files, \textit{IMAP} for mails, \textit{CalDAV} for calendar entries, and \textit{CardDAV} for contacts.
For web browsers, we use \textit{Web\-Extensions}\footnote{\url{https://wiki.mozilla.org/WebExtensions}}, which provide cross-browser functionality and an integration level similar to having an underlying protocol.
Applications are thus able to transparently operate on the knowledge graph (PIMO) managed by our app.
Especially in corporate scenarios, it is very convenient if users may just work with the resources in their contexts without caring whether they are actually spread across various sources like intranet shares, for example.

Utilizing only standard protocols has certain limitations due to their rather basic, low-level character.
Some activities, like writing a note or comment about a resource, can become inconvenient or non-intuitive.
To avoid this, we provide an additional sidebar as a single interaction point for using advanced features.
Users therefore do not need to learn a new (plugged-in) interface for each of their applications.
They can just keep using them the usual way having only the sidebar as a new UI to familiarize with.
From the development point of view, the effort of creating and maintaining plug-ins needed for higher level functionality is comparatively low to that of earlier SemDesks.
They can be realized as \textit{headless plug-ins} having very little functionality, often just the capability of \textit{sending out} in-app-events to the sidebar (that is why we also shortly call them "plug-outs").
In addition, their corresponding UI elements and logic are located in the sidebar, where they can be easily reused.
Plug-outs for different mail clients could, for example, share the same tagging UI.

\noindent
\textbf{Self-Reorganization.}
Features discussed so far primarily aim at our system's ease of use.
The other aspects mentioned in the beginning (scalability, missing "killer app") will be addressed using \textit{Managed Forgetting}, by which we understand an escalating set of measures: temporal hiding, condensation, adaptive synchronization, archiving and deletion of resources and parts of the knowledge graph \cite{forgetitbook}.
By having users work on cSpaces, we gather rich contextual information about all of their resources, which allows the system to semi-automatically help them in organizing their stuff.
Thus, cSpaces are continuously spawned, retracted, merged, condensed, or forgotten.
As an example, let us assume we do a consulting job for company XY.
The contract involves five meetings about different topics.
Our system could represent this by having an overall cSpace containing general information about XY, e.g. contact and contract information.
For each meeting, there could be an individual sub-cSpace about its respective topic.
Several months after the job has been completed, the system starts to remove details, e.g. train schedule to get to the meeting or auxiliary material for doing the presentation.
After some years have passed, the sub-cSpaces could be merged with their parent, since the separation into different meetings is not relevant anymore.
Only the most important items, e.g. individual reports or an overall final report, are kept.
All other items are either condensed, moved to an archive or deleted completely (which can be adjusted by the user on a general level).
An item's current and estimated future value for the user are therefore continuously assessed resulting in different forgetting measures like temporal hiding (e.g. some items during one of the meetings), deletion, etc.
This especially means that the system is able to reorganize itself to a certain extent, which especially includes a kind of tidying-up-itself functionality.
Some of the described features have already been implemented and successfully used in our research and industry prototypes \cite{manaforge, pimodiary}, however most of them are still under heavy development.\\[-0.3cm]

\noindent
\textbf{Demo.}
In an early proof-of-concept implementation based on \cite{maus2013weaving}, we already realized some of the file system, browser and calendar parts.
The screenshots in Figure \ref{fig_scr} show a typical feature of our system:
\vspace{-0.45cm}
\begin{figure}
\centering\includegraphics[width=1\textwidth]{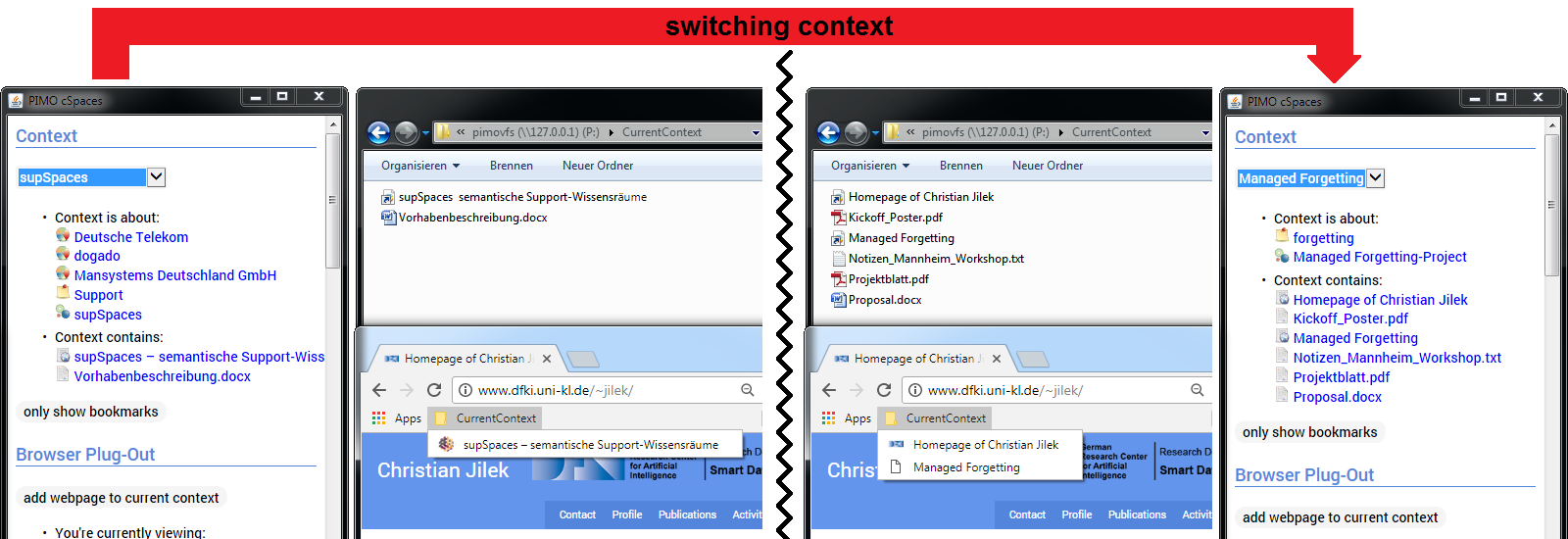}

\caption{
Screenshot showing sidebar, file explorer and browser before (left half) and after a context switch (right half), illustrating the effects of a dynamic reorganization of the system.
(Note: windows were rearranged for easier comparison.)
}
\label{fig_scr}
\end{figure}
\vspace{-0.45cm}
the user selects a different context using the sidebar.
As a consequence, the \textit{current context}, available as a folder in the file system as well as the browser, is dynamically reorganized by our app.
Note that the system tries to present meaningful views on the current context in each app: e.g., the view in the browser only contains web links.
To really get an impression of how the interaction with the system looks like, we kindly refer the reader to our online demo video\footnote{\url{https://pimo.opendfki.de/cSpaces/}}, which also shows some additional features.

\section{Conclusion \& Outlook}
In this paper, we presented a new SemDesk prototype based context spaces that users directly interact with and work on.
The system is transparently integrated using mostly standard protocols complemented by a sidebar for advanced features.
Users may thus stick to their favorite applications which should strongly contribute to the overall ease of use.
Learning efforts are presumably low due to the sidebar being the only new UI that is introduced.
By exploiting its collected context information and applying features of Managed Forgetting, the system is able to dynamically reorganize itself which also includes a kind of tidying-up-itself functionality.
We therefore expect it to be more scalable than its predecessors while providing new levels of user support.

Nevertheless, a lot of functionality still needs to be fully implemented and evaluated.
We plan to do extensive user studies once the system matures.\\

\noindent
\textbf{Acknowledgements.}
This work was funded by the Deutsche Forschungsgemeinschaft (DFG, German Research Foundation) -- DE 420/19-1.

\bibliographystyle{splncs03}
\bibliography{paper}

\end{document}